\documentclass[letterpaper, 10 pt, conference]{ieeeconf}

\IEEEoverridecommandlockouts
\usepackage{cite}

\usepackage[pagewise]{lineno}
\usepackage{amsmath}
\usepackage{amssymb}
\usepackage{wrapfig}
\usepackage{pifont}
\usepackage{tabularray}

\usepackage{multirow} 
\usepackage{hyperref}
\usepackage{mathtools}
\usepackage{bm}
\usepackage{algorithm}
\usepackage{algpseudocode}
\algnewcommand{\Initialize}[1]{%
  \State \textbf{Initialize:}
  \Statex \hspace*{\algorithmicindent}\parbox[t]{.8\linewidth}{\raggedright #1}
}
\usepackage{graphicx}
\usepackage{subcaption}
\usepackage{xcolor}
\usepackage{booktabs}

\graphicspath{{./figures/}}

\title{\LARGE \bf
Agile Mobility with Rapid Online Adaptation \\ via Meta-learning and Uncertainty-aware MPPI
}
\author{Dvij Kalaria, Haoru Xue, Wenli Xiao, Tony Tao, Guanya Shi, and John M. Dolan
\thanks{The authors are with the Robotics Institute, Carnegie Mellon University {\tt\small \{dkalaria, haorux, wxiao2, longtao, guanyas, jdolan\}@andrew.cmu.edu}}}
\begin{document}
%\linenumbers % Uncomment this to enable line numbers in the peer review

\maketitle
\thispagestyle{empty}
\pagestyle{empty}

%%%%%%%%%%%%%%%%%%%%%%%%%%%%%%%%%%%%%%%%%%%%%%%%%%%%%%%%%%%%%%%%%%%%%%%%%%%%%%%%
\begin{abstract}
    Modern non-linear model-based controllers require an accurate physics model and model parameters to be able to control mobile robots at their limits. Also, due to surface slipping at high speeds, the friction parameters may continually change (like tire degradation in autonomous racing), and the controller may need to adapt rapidly. Many works derive a task-specific robot model with a parameter adaptation scheme that works well for the task but requires a lot of effort and tuning for each platform and task. In this work, we design a full model-learning-based controller based on meta pre-training that can very quickly adapt using few-shot dynamics data to any wheel-based robot with any model parameters, while also reasoning about model uncertainty. We demonstrate our results in small-scale numeric simulation, the large-scale Unity simulator, and on a medium-scale hardware platform with a wide range of settings. We show that our results are comparable to domain-specific well-engineered controllers, and have excellent generalization performance across all scenarios.
\end{abstract}

\vspace{-5mm}
\section{Introduction}
\vspace{-1mm}

Recently, there has been a surge in robot controllers that enable mobile robots to operate at their limits, to the extent that they can possibly beat humans in competitive sports like drone racing \cite{Kaufmann2023ChampionlevelDR} and autonomous car racing \cite{Wischnewski2022IndyAC}, spurred by recent competitions like the Indy Autonomous Challenge, A2RL, F1Tenth, etc. However, we are faced with various practical challenges in designing autonomous controllers for such tasks, due to: 1) delays in computation and system leading to inaccurate controls; 2) sensor noise; 3) imperfect model; 4) a dynamic environment due to changing surface, tire temperature, and weather conditions; 5) safety concerns.

\begin{figure}
    \centering
    \includegraphics[width=.47\textwidth]{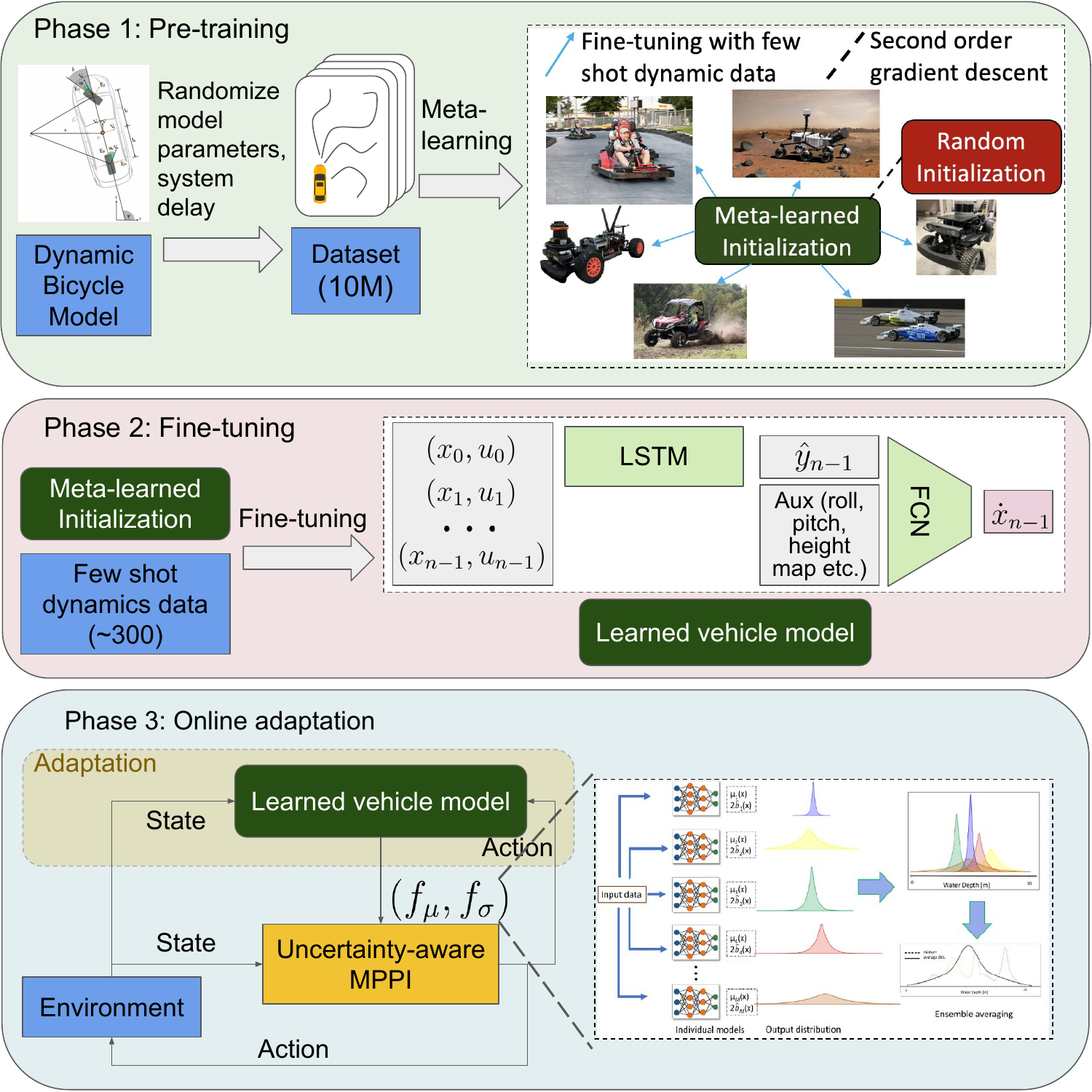} 
    \caption{Overview of our approach. \textbf{Phase 1} involves pre-training a model initialization from numeric sim data ($\sim10$M) which can quickly adapt to any vehicle platform with few-shot dynamic data. \textbf{Phase 2} involves fine-tuning the model from very few dynamic data ($\sim300$) on the platform. \textbf{Phase 3} involves deploying the learned model online for control using uncertainty-aware MPPI while adapting the model in real-time from online dynamic data.}
    \label{fig:overview}
    \vspace{-6mm}
\end{figure}

One way of tackling this problem is to derive a simplified physics model and adapt the model parameters. But this requires significant engineering effort to design such a scheme, especially for autonomous racing, as its hardware model depends on many model parameters as well as system delays which can't all be easily modeled for all systems. In this work, we aim towards an approach that can be generalized to any vehicle platform including system delay. We devise a meta learning approach where we train an easily adaptable end-to-end sequence prediction model that takes the history of states and actions as input and outputs the state derivative, which is integrated to obtain the next state. Learning a vehicle model from online dynamics data often leads to local overfitting \cite{Senanayake2024TheRO}. We propose using uncertainty-aware cost for Model Predictive Path Integral (MPPI) control, which discourages taking actions that the learned model is not confident about. An overview of our approach is given in Figure \ref{fig:overview}. In conclusion, we make the following contributions:
\begin{itemize}
    \item We design a meta-learning-based initialization for an end-to-end vehicle model that can quickly adapt to any new vehicle with very little dynamics data and can adapt to changing parameters
    \item We use an uncertainty-aware MPPI that reasons about epistemic uncertainty coming from local over-fitting during online model adaptation
    \item We test our approach in numeric simulation, hardware, and in our custom-designed Unity game engine-based simulation in various scenarios. We open-source our simulator and code at: \href{https://github.com/DRIVE-LAB-CMU/meta-learning-model-adaptation}{https://github.com/DRIVE-LAB-CMU/meta-learning-model-adaptation}
\end{itemize}

The rest of the paper is structured as follows: Section \ref{sec:c5_previous_works} describes related work. Section \ref{sec:c5_method} describes the methodology in detail.  Section \ref{sec:c5_results} describes some results in various simulators and on a real-world RC-car. Section \ref{sec:c5_conclusion} concludes our discussion and prepares the ground for future work.

\section{Previous work} \label{sec:c5_previous_works}

In recent times, there have been numerous works on model-learning-based control across different domains. Many of these works aim towards directly learning a policy that can perform a certain task or execute a skill. Some of these works directly deploy the policy on hardware that is trained in a simulator with domain randomization to make the policy robust against model parameter errors \cite{Chen2021UnderstandingDR}. Some works employ sim-to-real techniques that modify the training objective to be able to predict the model parameters first from the history and then use the parameters to condition the policy \cite{Kumar2021RMARM}. These approaches work well if the actual parameters are within the training distribution in the simulator. 

If the parameters drift significantly, which is common in racing due to tire degradation, these approaches fail to generalize to those scenarios. To address this challenge, there is also a need to adapt the parameters in a closed loop based on feedback from the environment. This is addressed in \cite{Xiao2023SafeDP}. However, these approaches are still trained to perform a specific task. To train on a different task, large-scale training must be carried out with the same procedure. 

There is another body of work which aims towards learning model parameters and then using them to simulate the learned model. The learned model is used to formulate the optimal control problem using MPC \cite{Liniger2015OptimizationbasedAR} or MPPI \cite{Williams2016AggressiveDW}. Using MPPI has become more popular, as it allows highly non-linear models. Most works like \cite{Kumar2021RMARM} use an end-to-end model without a specific model structure, while some have a specific structure like \cite{Chrosniak2023DeepDV} and \cite{Kalaria2023AdaptivePA}. A complete end-to-end model makes it flexible even if the model used to represent the system dynamics is not accurate, as it does not assume any structure. However, this model is prone to over-fitting to a specific scenario if trained directly online, as online data may not have enough diversity. \cite{Nagabandi2018LearningTA} proposes using meta-reinforcement learning (meta-RL) for model initialization that can adapt to changes in the environment for model-based RL tasks such as one of the motor controllers becoming dysfunctional for a quadruped. They are able to show that using meta-learning, the trained model-based RL policy is robust to such model changes. However, our aim is to deploy the controller directly on hardware for any task specification using MPPI. The task in MPPI can be defined by defining the cost function. Using the trained model directly for agile tasks like high-speed driving with a vehicle operating at its control limits may lead to over-fitting over the locally seen state space, leading to the vehicle losing control. Towards this end, we propose an uncertainty-aware approach to systematically learn to perform well on states it is less confident on, as well as be able to prefer assured safe actions rather than focus on states that might offer better optimality but poor confidence. 

\vspace{-2mm}
\section{Methodology} \label{sec:c5_method}

\vspace{-1mm}

\subsection{Dynamic bicycle model} \label{sec:dynamic_model}
\vspace{-1mm}

We will first describe the most popular vehicle model used in the literature to control a racecar \cite{Kabzan2019AMZDT,Kalaria2022DelayAwareRC,Kalaria2021LocalNO,Liniger2015OptimizationbasedAR}, the dynamic bicycle model. We will use this model to generate data for pre-training (see Section \ref{sec:meta_learning}). The dynamic model has states $p_x$, $p_y$, and $\phi$ as position and orientation in the global frame; longitudinal velocity $v_x$, lateral velocity $v_y$, and yaw angular velocity $\omega$ in the vehicle's body frame. For the dynamic model, $F_{r,x}$ is the longitudinal force on the rear tire in the tire frame assuming a rear-driven vehicle, $F_{f,y}$ and $F_{r,y}$ are the forces on the front and rear tires, respectively, and $\alpha_f$ and $\alpha_r$ are the corresponding slip angles.
% which are assumed to be numerically less in magnitude such that $\cos(p) \approx 1$ and $\cos(r) \approx 1$ \qin{can we ignore them?}. 
We denote the mass of the vehicle $m$, the moment of inertia in the vertical direction about the center of mass of the vehicle $I_z$, the length of the vehicle from the COM (center of mass) to the front wheel $l_f$, and the length from the COM to the rear wheel $l_r$. $B_{f/r}$, $C_{f/r}$, $D_{f/r}$ are the Pacejka tire model parameters \cite{Pacejka1982TireAV} specific to the tire and track surface. $\delta$ is the steering command. $C_{m1}$ and $C_{m2}$ are longitudinal parameters that dictate the longitudinal force generated by the vehicle given the throttle command, $d$. $C_{lf}$ is the kinetic friction force and $C_d$ is the aerodynamic drag coefficient.

%The dynamic model is complex and takes into consideration the lateral tire slip but requires time-intensive identifications to determine the Pacejka tire model parameters of the tire, drivetrain, and aerodynamic resistance parameters. It is highly suitable for autonomous racing, as it accurately models the dynamics at cornering speeds during turns. However, accurate values of parameters are needed, and these values may change over time. %which may threaten the accuracy of the model. 
%Also, the tire lateral slip curves must be re-calibrated if the surface changes, which is common for autonomous racing. Here, 

% \qin{(any definition for slip curves?)} : Added above, shall I add here again?

% EQN here
\vspace{-4mm}

\begin{equation} \label{eqn:dyn_model}
\begin{split}
&\begin{bmatrix}
    \dot{p}_x \\
    \dot{p}_y \\
    \dot{\phi} \\
    \dot{v}_x \\
    \dot{v}_y \\
    \dot{\omega} \\
\end{bmatrix}
= \begin{bmatrix}
    v_x \cos (\phi) - v_y \sin (\phi)\\
    v_x \sin (\phi) + v_y \cos (\phi) \\
    \omega \\
    \frac{1}{m} (F_{r,x} - F_{f,y} \sin(\delta) + m v_y \omega)\\
    \frac{1}{m} (F_{r,y} + F_{f,y} \cos(\delta) - m v_x \omega)\\
    \frac{1}{I_z} (F_{f,y} l_f \cos(\delta) - F_{r,y} l_r) \\
\end{bmatrix} \\ 
\end{split}
\end{equation}
% \vspace{-2mm}
% \begin{subequations}
% \begin{align}
% &\text{where } F_{r,x} = (C_{m1} - C_{m2} v_x) d - C_r - C_d v_x^2 \\
% &F_{f,y} = D_f \sin(C_f \tan^{-1}(B_f \alpha_f)), \alpha_f = \delta - \tan^{-1}\left(\frac{\omega l_f + v_y}{v_x}\right) \\
% &F_{r,y} = D_r \sin(C_r \tan^{-1}(B_r \alpha_r)), \alpha_r = \tan^{-1}\left(\frac{\omega l_r - v_y}{v_x}\right)
% \end{align}
% \end{subequations}
where $F_{r,x} = (C_{m1} - C_{m2} v_x) d - C_{lf} - C_d v_x^2$, $F_{f,y} = D_f \sin(C_f \tan^{-1}(B_f \alpha_f)), \alpha_f = \delta - \tan^{-1}\left(\frac{\omega l_f + v_y}{v_x}\right)$, and $F_{r,y} = D_r \sin(C_r \tan^{-1}(B_r \alpha_r)), \alpha_r = \tan^{-1}\left(\frac{\omega l_r - v_y}{v_x}\right)$.

% \qin{$C_{m1}$, $C_{m2}$, $C_d$ are...} : Added above

\vspace{-1mm}
\subsection{Meta learning} \label{sec:meta_learning}
\vspace{-1mm}

We now formulate our problem as a meta-learning problem. Let us define the state as $x$ and an action $u$ as described in Section \ref{sec:dynamic_model}. Our trainable model $f(.)$ takes the history of states and actions $H_i = \{(x_{i-n},u_{i-n}), (x_{i-n+1},u_{i-n+1})..., (x_{i},u_{i})\}$ and outputs the next state, $x_{i+1}$. The network outputs $y_i = \frac{x_{i+1}-x_i}{\Delta t}$. For each new environment $r$, the optimal weights representing the environment are denoted as $f_r$. We use the model-agnostic meta learning \cite{Finn2017ModelAgnosticMF} algorithm to learn weights which, if initialized with, can quickly adapt from online data to a new scenario in $K$ steps. For this we need second-order derivatives, as described in algorithm \ref{alg:maml}. For training, we generate a wide range of scenarios as described in Section \ref{sec:data_gen} as separate tasks. 
A task $r$ denoted by $\tau_r$ consists of data sample pair $(X_r, Y_r)$ with size $N$ 
where $X_r = \{ H_1, H_2 ... H_{N}\}$ and $Y_r = \{ y_1, y_2 ... y_{N}\}$. The loss function is defined as the mean square prediction error as follows:- $\text{Loss}(f(\theta, X),Y) = \frac{1}{M} \sum_{i=0}^{M-1} (f(\theta, X(i \Delta t)) - Y(i \Delta t))^2$ where $\theta$ is the model parameters, $M$ is the horizon length.

% FIG HERE
% \begin{figure}
%     \centering
%     \includegraphics[width=.35\textwidth]{figures/meta-learning.png}
%     \caption{Meta-learning}
%     \label{fig:meta-learning}
% \end{figure}

% ALGORITHM HERE
% \vspace{3mm}
\begin{algorithm}[htbp]
\caption{Model-Agnostic Meta-Learning (MAML)}
\label{alg:maml}
\begin{algorithmic}
\Initialize{$\theta = \text{initialize parameters}$}

\For{each episode $t$}
  \State $\tau_r \sim \mathcal{T}$ // Sample a task $\tau_r$ from batch $\mathcal{T}$

  \State $\theta_r^1 = \theta$ // Initialize inner loop parameters

  \For{$k = 1$ \text{to} $K$}
    \State $\hat{Y}_r^k = f(\theta_r^k, X_r)$ // Evaluate model on task data
    \State $\ell_r^k = \text{Loss}(\hat{Y}_r^k, Y_r)$ // Calculate loss
    \State $\theta_i^{k+1} = \theta_i^k - \alpha \nabla_{\theta_i^k} \ell_i^k$ // Update inner loop parameters
  \EndFor

  \State $\mathcal{L} = \frac{1}{|\mathcal{T}|} \sum_{\tau_i \in \mathcal{T}} \text{Loss}(f(\theta_r^K, X_r), Y_r)$ // Calculate meta-loss
  \State $\theta \leftarrow \theta - \beta \nabla_{\theta} \mathcal{L}$ // Update model parameters
\EndFor
\end{algorithmic}
% \vspace{-3mm}
\end{algorithm}

\vspace{-1mm}
\subsection{Dataset generation} \label{sec:data_gen}
\vspace{-1mm}

We use the dynamic model from Section \ref{sec:dynamic_model} to generate a dataset for meta-learning. We randomize all model parameters including $m, I_z, C_{f/r}, D_{f/r}, B_{f/r}, l_f, l_r,C_{m1},C_{m2}, C_{lf}, C_d$. We take $\delta = K_d \delta_{\text{cmd}} + K_{\text{bias}}$ where $K_d$ and $K_\text{bias}$ are constants so that the learned initialization can also pick up steering biases, common in real systems. We generate random but smooth action commands over time length $t_h$ as follows using a Fourier series function with random coefficients:

% EQN HERE
\begin{equation}
    u(t) = C_0 + \sum_{k=1}^{k=K-1} C_k \sin \left (\frac{2 \pi t}{k \Delta t} \right ) 
\end{equation}
where $C_0, C_1 \cdots C_{K-1}$ are randomly sampled coefficient vectors of length $2$ which dictate the shape and smoothness of the action-generating function, $u(t)$. The coefficient vectors $C_0, C_1 ... C_{K-1}$ are such that $\Sigma_{i=0}^{i=K-1} C_i = [ 1 \ 1]^T$, which ensures that the generated control commands are within the $[-1,1]$ range. We generate state sequence $x(t)$ using the differential equations in Eqn. \ref{eqn:dyn_model}. To make the initialization robust to delays in the system, we then shift the generated state sequence by system delay time $t_d$, which is also randomly generated. The generated state sequence $x(t)$ is for times $t=[t_d, t_d+t_h)$ and the action sequence is from $t=[0,t_h)$. We use states and actions from $t=[t_d,t_h)$ to get the data sample pair $(X_r, Y_r)$ for task $r$, which is used for meta-learning as described in Section \ref{sec:meta_learning}. As the Fourier series can represent any function in theory, the sampled continuous action sequence should theoretically be able to represent any action sequence pattern. Hence, the trained meta-learned model from these data should be able to adapt to any given dynamic data sequence. It is important to note that although we generate data with the dynamic bicycle model, the meta-learned model can still generalize to other vehicle types that do not follow Eqn. \ref{eqn:dyn_model}. In Section \ref{sec:c5_results}, we will see how our model can generalize to other vehicle models like a go-kart with fixed rear axle, which cannot be described by Eqn. \ref{eqn:dyn_model}; scenarios with high longitudinal slip, which is assumed to be zero in Eqn. \ref{eqn:dyn_model}; or cars with large width and small turn radius, for which the bicycle assumption is inaccurate.

\subsection{Model architecture}

Because the next state prediction task is sequential in nature, we use an LSTM encoder. Then we use a fully connected network to output $\dot{v_x}, \dot{v_y}, \dot{\omega}$. These derivatives are assumed to be constant for $\Delta t$. We also append auxiliary information of size $L$ to the encoded history, which may include features like roll, pitch, longitudinal slip, etc. that may be relevant for prediction but not included in dynamic state. This is used for some scenarios in Section \ref{sec:c5_results}. 

% \begin{figure}
%     \centering
%     \includegraphics[width=0.5\textwidth]{c5/model_arch.png}
%     \caption{Model architecture}
%     \label{fig:c5_model_arch}
% \end{figure}

\subsection{Uncertainty estimation} \label{sec:c5_uncert_est}

For a safety-critical problem, especially for autonomous racing, where safety is also closely tied to the performance, it may also be necessary to quantify if the trained model is confident about the prediction to avoid the vehicle losing control at its limits. We use ensemble-based epistemic uncertainty prediction, where we randomly initialize weights $f^{(0)},f^{(1)}...,f^{(l-1)}$ and train all of them along the same objective. Now when we make a prediction, we take the mean of all predictions as $\mu_y$ and variance $\sigma_y$. We define the functions $f_mu$ and $f_\sigma$ accordingly:

\begin{equation}
\begin{aligned}
    &\mu_y = f_\mu(x) = \frac{1}{l} \sum_i f^{(i)} (x) \\
    &\sigma_y = f_\sigma(x) = \frac{1}{l} \sum_i (f^{(i)} (x)-\mu_y)^2
\end{aligned}
\end{equation}

Since all randomly initialized models are trained on the same objective, they should produce the same output, $y$, for inputs within the training distribution. However, for inputs outside the training distribution, the models will likely generate different outputs, leading to increased output variance, $\sigma_y$, due to their lack of prior exposure to such inputs. We leverage this I.I.D. property of training to estimate whether a given input $H$ lies outside the training distribution.

\subsection{Model Predictive Path Integral (MPPI) Control}

MPPI is popular for control when the model is highly non-linear, as in our case. Consider our non-linear dynamical system: 
% EQN HERE
\begin{equation}
    x_{t+1} = f(x_i,u_i)
\end{equation}
where $x_k \in \mathcal{R}^{n_x}$ and $u_k \in \mathcal{R}^{n_u}$ is the control input. With the control horizon length $K$, we can denote a trajectory of states, $\mathbf{x} = [x_0^T,...,x_{K-1}^T]^T$, the mean control sequence $\mathbf{v} = [v_0^T,...,v_{K-1}^T]^T$ and the injected Gaussian noise $\mathbf{\epsilon} = [\epsilon_0^T,...,\epsilon_{K-1}^T]$. The control sequence $\mathbf{u} = \mathbf{v} + \mathbf{\epsilon}$. The original MPPI problem aims to solve the following problem:
% EQN HERE
\begin{equation}
\begin{aligned}
& \min _{\mathbf{v}} J(\mathbf{v})= \mathbf{E}\left[\sum_{k=0}^{K-1}\left(q\left(x_{k}\right)+\frac{\lambda}{2} v_{k}^{\top} \Sigma_{\epsilon}^{-1} v_{k}\right)\right] 
\end{aligned}
\end{equation}
where $\lambda$ is a hyperparameter, and $x(t)$ is the measured current state at time $t$. Note that the state-dependent cost $q(x_k)$ can take any form, which can be set accordingly based on the task. In our case, we will mostly look at the task of tracking, hence $q(x_k)$ is defined as follows:
% EQN HERE
\vspace{-1mm}
\begin{equation}
    q(x_k) = (x_k - x_{\text{ref},k})^2
\vspace{-1mm}
\end{equation}
% \vspace{-1mm}
where $x_{\text{ref},k}$ is the reference state at time $t+k\Delta t$, which needs to be tracked as closely as possible. The MPPI controller is derived by minimizing the KL-divergence between the current controlled trajectory distribution and the optimal distribution (as in \cite{Williams2018RobustSB}) to solve the problem. We sample a large number of simulation trajectories using our learned model $f$. The cost of the $m^\text{th}$ sample trajectory, $S_m$ is defined as follows:
% EQN HERE
\begin{equation*}
S_{m}=\sum_{k=0}^{K-1} q\left(x_{k}^{m}\right)+\gamma\left(v_{k}\right)^{\top} \Sigma_{\epsilon}^{-1}\left(v_{k}^{m}+\epsilon_{k}^{m}\right)
\end{equation*}

The MPPI algorithm calculates the optimal control $v^+$ by taking the weighted average of all sample trajectories as follows: 
% EQN HERE
\begin{equation*}
\mathbf{v}^{+}=\sum_{m=1}^{M} \omega_{m} \mathbf{u}^{m} / \sum_{m=1}^{M} \omega_{m} \\
\omega_{m}=\exp \left(-\frac{1}{\lambda}\left(S_{m}-\beta\right)\right)
\end{equation*}
where $\beta = \min_{m=1,...,M} S_m$ is the smallest cost among $m$ sampled trajectories and it is used to avoid numerical overflow while keeping the same solution. For the next control cycle, the mean control, $v_{t+1}$ is set to the previous command, $v_t^+$, but shifted by $1$ as suggested in \cite{Yi2024CoVOMPCTA}, i.e., $v_{t+1,k} = v_{t,k+1}^+ \text{ for } k \in \{0,1,..,K-2\}$ as we apply the first control command in each iteration in receding horizon fashion. For the last step, we set $v_{t+1,K-1}=v_{t+1,K-1}^+$.

\subsection{Uncertainty-aware MPPI}

We do not want to prioritize actions where the model is not confident. There are multiple approaches to tackle this, one of which is as given in \cite{Yin2022RiskAwareMP}, where the worst-case cost trajectory of a sample of trajectories sampled from each action sample with uncertainty is chosen. But to practically implement this, we need to sample additional trajectories to get the worst case sample for each sampled action sequence. We however adopt a cost-efficient approach by adding uncertainty variance as calculated in Section \ref{sec:c5_uncert_est} of each sample to discourage choosing actions with high uncertainty. The net cost after adding this additional cost with $\gamma$ scaling factor is:
% EQN HERE
\begin{equation}
    q(x_k) = \gamma \sigma_k + (x_k - x_{\text{ref},k})^2
\end{equation}

\subsection{Online adaptation}

Once we have the control algorithm in place, it is necessary to adapt the model to new conditions. The pre-trained model that we got from Section \ref{sec:meta_learning} needs to be fine-tuned on the on-policy data. For this we first need a good enough model, i.e., one that gives some meaningful predictions. For this, we use a simple controller at the beginning to adapt the pre-trained model on the given environment with some few-shot data. Then, we switch to use the trained model for the task and continue adapting the model with on-policy data. In a continuously evolving environment like that in autonomous racing, fast adaptation is also  necessary. We use simple gradient descent to fine-tune our model, where we keep the last $b$ state-action pairs in buffer $B = \{ (x_{t-b+1},u_{t-b+1}) , (x_{t-b+2},u_{t-b+2}) ..., (x_{t},u_{t})\}$ and use them to take the gradient descent step as follows:
% EQN HERE
\begin{equation}
\begin{aligned}
    &L = \sum_{i=0}^{i=b-n-1} \text{Loss}(f(\theta, X_i), Y_i) \\
    &\theta^+ = \theta - \lambda \frac{\partial L}{\partial \theta}
\end{aligned}
\end{equation}
Here, Loss and $(X_i,Y_i)$ are defined the same as in Section \ref{sec:meta_learning}. The training should be fast, as it needs to be performed while the car is running in real-time, hence we need to choose the buffer size $B$ accordingly.

\section{Experiment} \label{sec:c5_results}

We now present our results on various platforms and compare with the following baselines: \textbf{a)} No model adaptation, incorrect vehicle parameters. \textbf{b)} No model adaptation, correct model parameters, assume no delay. This is to show how delay, which is not trivial to estimate, plays an important role in designing an accurate control at limiting speeds. \textbf{c)} Our proposed approach but with random instead of meta-learned initialization. \textbf{d)} Proposed approach without considering uncertainty. \textbf{e)} Proposed approach but with an initialization learned with $K=0$, i.e., no meta-learning. \textbf{f)} Proposed approach. \textbf{g)} Assuming dynamic bicycle model and adapting model parameters (same as APACRace \cite{Kalaria2023AdaptivePA}).

We aim to answer the following questions through the experiments: \textbf{Q1}) How does our proposed approach perform in numeric simulation with the same model parameter distribution as the data on which the meta-learning is trained? and How does its tracking performance compare with other baseline approaches? \textbf{Q2}) How does our performance compare to other baselines on out-of-distribution scenarios like vehicle models which cannot be represented by Eqn \ref{eqn:dyn_model}? \textbf{Q3}) Can we deploy our approach in real time on hardware and how does our approach perform in comparison to baselines? All videos of results can be viewed at \href{https://sites.google.com/view/meta-learning-model-adaptation}{https://sites.google.com/view/meta-learning-model-adaptation}.

\subsection{Numeric simulation} \label{sec:numeric_sim}

We first evaluate our approach in numeric simulation. The dynamics for the numeric simulation are defined as given in Section \ref{sec:dynamic_model}. We evaluate our results on an oval trajectory, which is a good platform to test drifting in one direction for any control algorithm. We do $10$ rollouts for each approach with randomly chosen model parameters with the same distribution used for training in Section \ref{sec:meta_learning} to see how each approach compares in terms of tracking. 

\begin{figure}
% \vspace{-5mm}
\centering
\begin{subfigure}{.18\textwidth}
    \centering
    \includegraphics[width=\textwidth]{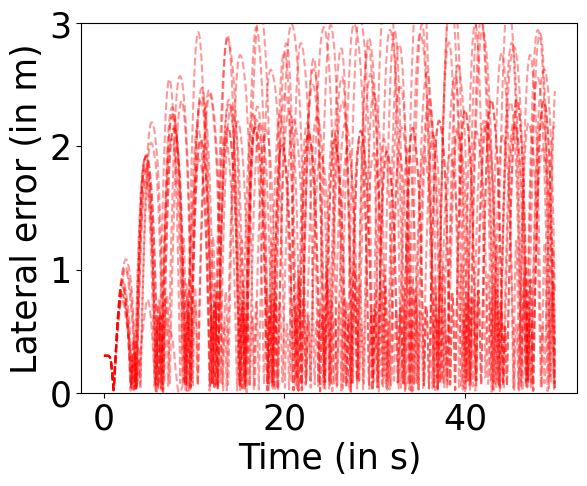}
    \caption{}
    % \caption{With no adaptation using incorrect model params}
    \label{fig:c5_no_adaptation_}
\end{subfigure}
\begin{subfigure}{.18\textwidth}
    \centering
    \includegraphics[width=\textwidth]{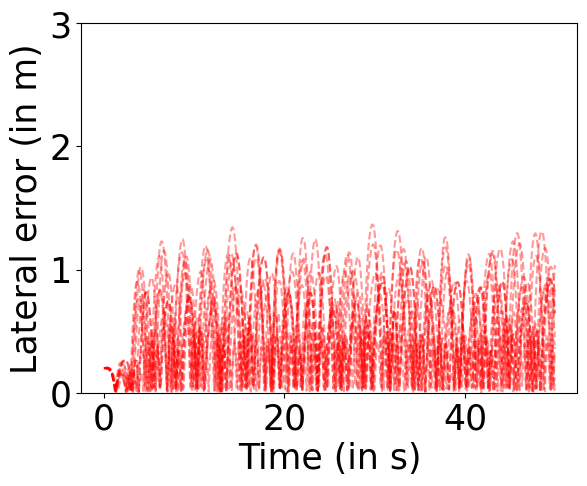}
    \caption{}
    % \caption{With no adaptation using correct params but assuming no delay}
    \label{fig:c5_no_adaptation_incorrect_delay_}
\end{subfigure}
\begin{subfigure}{.18\textwidth}
    \centering
    \includegraphics[width=\textwidth]{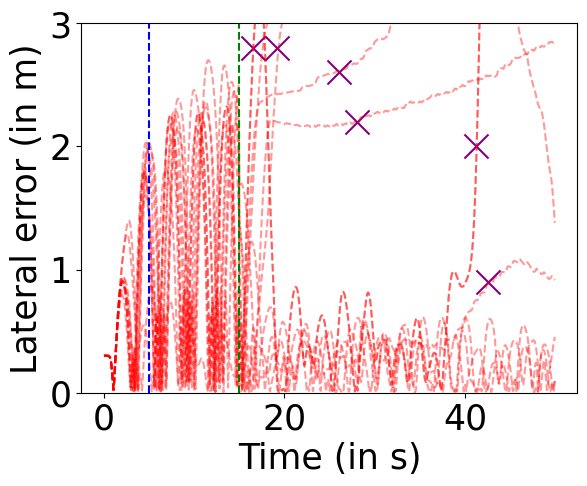}
    \caption{}
    % \caption{With adaptation starting with random initialization}
    \label{fig:c5_with_adaptation_random_init_}
\end{subfigure}
\begin{subfigure}{.18\textwidth}
    \centering
    \includegraphics[width=\textwidth]{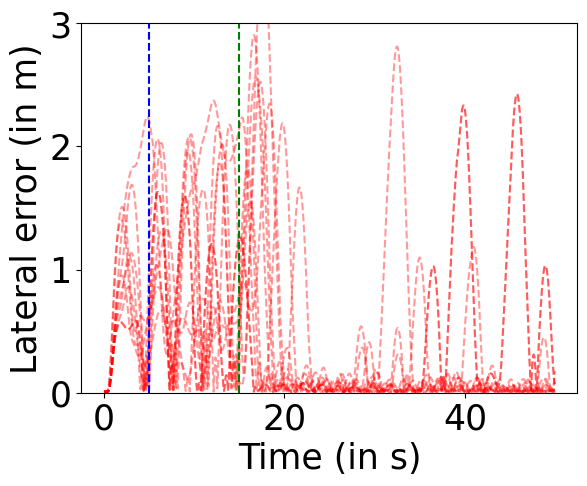}
    \caption{}
    % \caption{With adaptation with pre-training initialization no uncertainty estimation}
    \label{fig:c5_with_adaptation_no_uncertainty_}
\end{subfigure}
\begin{subfigure}{.18\textwidth}
    \centering
    \includegraphics[width=\textwidth]{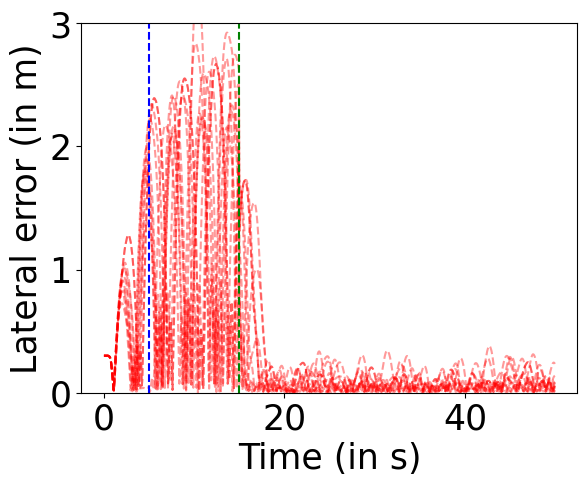}
    \caption{}
    % \caption{With adaptation with average initialization with uncertainty estimation}
    \label{fig:c5_with_avg_adaptation_with_uncertainty_}
\end{subfigure}
\begin{subfigure}{.18\textwidth}
    \centering
    \includegraphics[width=\textwidth]{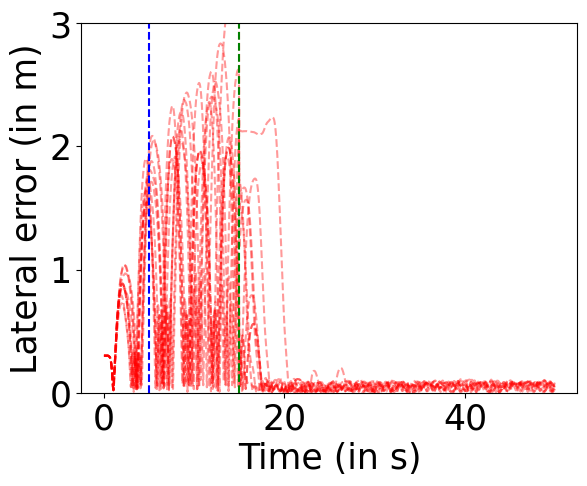}
    \caption{}
    % \caption{With adaptation with pre-training initialization with uncertainty estimation}
    \label{fig:c5_with_adaptation_with_uncertainty_}
\end{subfigure}
\begin{subfigure}{.18\textwidth}
    \centering
    \includegraphics[width=\textwidth]{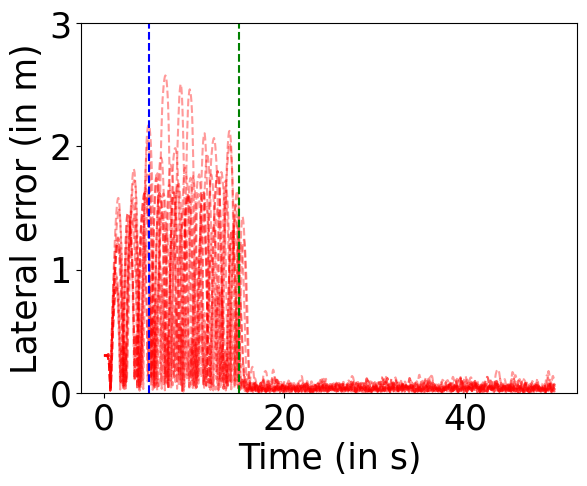}
    \caption{}
    % \caption{APACRace}
    \label{fig:c5_with_APACRace_}
\end{subfigure}
\begin{subfigure}{.25\textwidth}
    \centering
    \includegraphics[width=\textwidth]{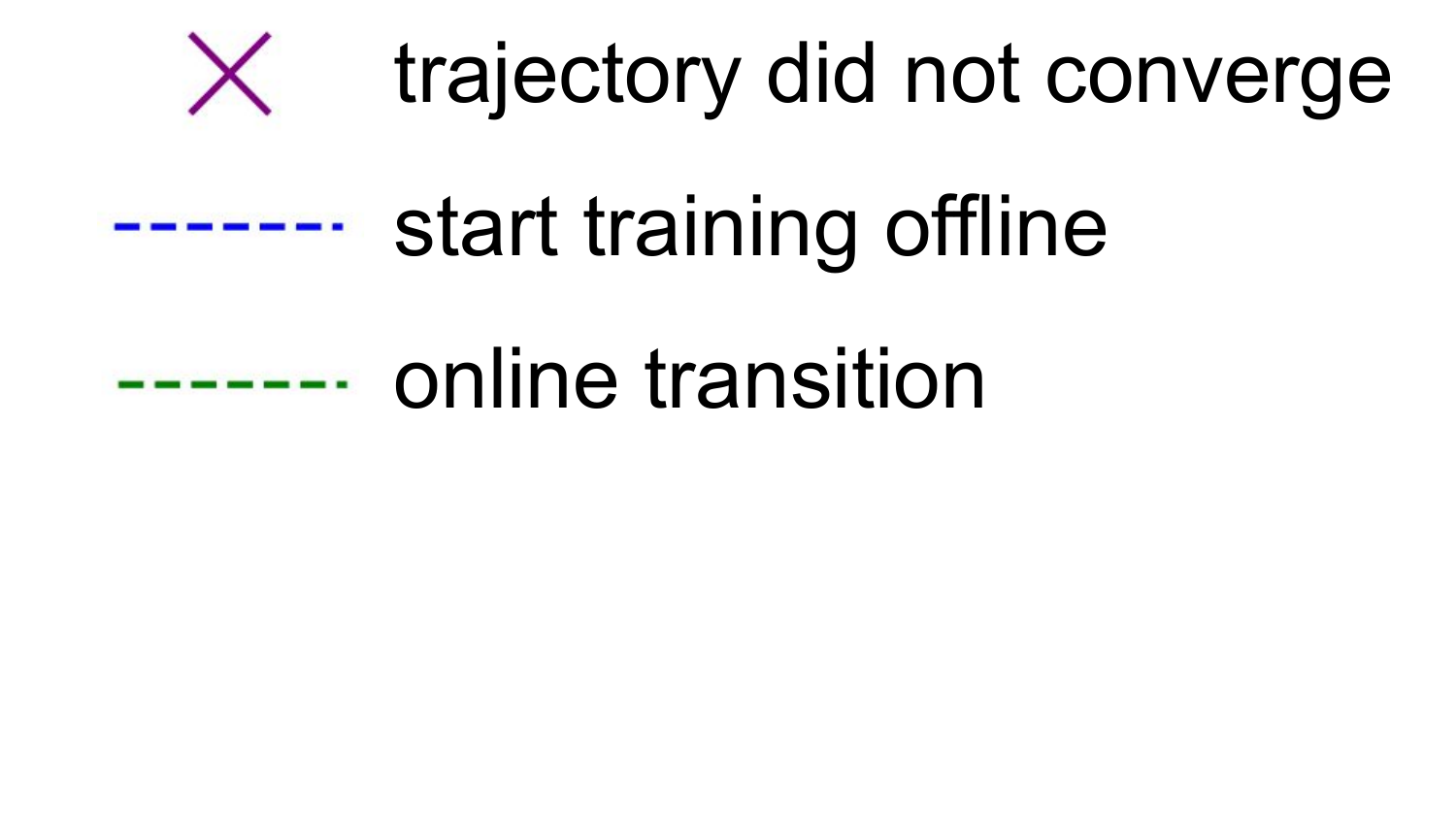}
    \label{fig:legend}
\end{subfigure}
\caption{Lateral errors for numeric sim experiment for $10$ environments with random model parameters. Please refer to Section \ref{sec:c5_results} for definitions of baselines a-g.}
\label{fig:c5_lat_errs}
\vspace{-5mm}
\end{figure}

\begin{figure}
\centering
\begin{subfigure}{.23\textwidth}
    \centering
    \includegraphics[width=\textwidth]{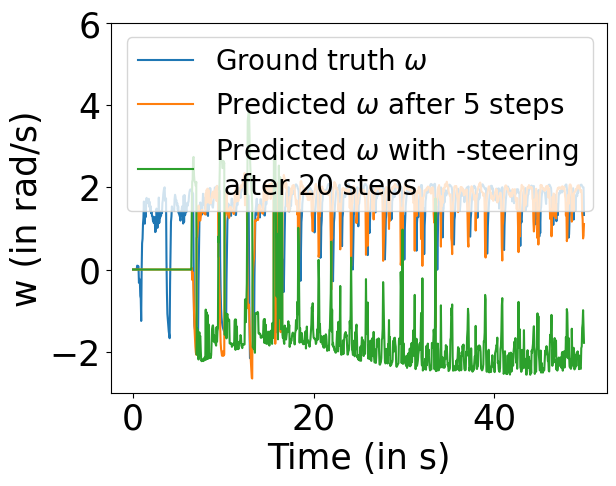}
    \caption{with meta-learned initialization}
    \label{fig:ws_pre_training}
\end{subfigure}
\begin{subfigure}{.23\textwidth}
    \centering
    \includegraphics[width=\textwidth]{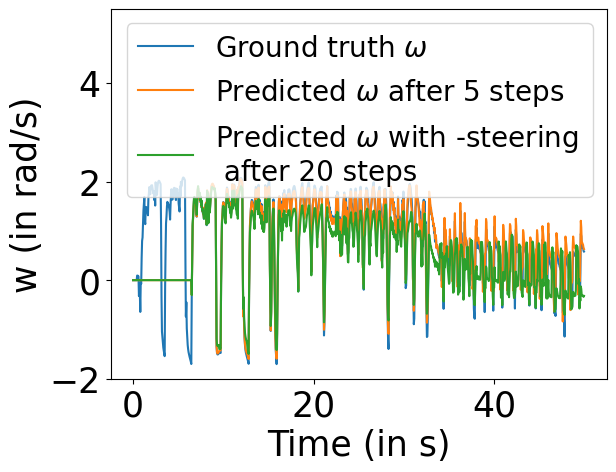}
    \caption{with random initialization}
    \label{fig:ws_without_pre_training}
\end{subfigure}
\caption{Predicted $\omega$ comparison for numeric sim experiment}
\label{fig:c5_ws}
\end{figure}

The lateral error plots for all approaches are given in Figure \ref{fig:c5_lat_errs}. As can be seen in Figure \ref{fig:c5_with_adaptation_no_uncertainty_}, where our proposed approach is used without uncertainty estimation, the lateral error suddenly jumps at some points. This occurs because, at those points, the model overfits due to the exclusive reliance on on-policy data from left turns. Due to this, at some points it mistakes right steers as potential commands for high angular velocities and thus ends up executing them, leading to sudden jumps in lateral error. If pre-training is not used, as seen in Figure \ref{fig:c5_with_adaptation_random_init_} it takes a long time to settle for a well-trained model that can track the trajectory, while with pre-training, it very rapidly adapts to the new environment and brings down the lateral error immediately, as seen in Figure \ref{fig:c5_with_adaptation_with_uncertainty_}. To understand more clearly how pre-training helps, we present a predicted angular velocity plot in Figure \ref{fig:c5_ws}. We compare the angular velocities achieved if the same steering thaw applied at that time step is applied further for $10$ more time steps and if the negative of that steering is applied for the next $10$ time steps. If the model has learned correctly: due to the model's inherent symmetry assuming no steering bias, it should be able to predict opposite angular velocities for opposite steering if applied for a sufficient number of time steps. If it does, it means it has understood that the steering is symmetrical on both sides. As seen in Figure \ref{fig:ws_pre_training}, it predicts opposite angular velocities for left and right steering right from the beginning, it just adjusts the rate magnitude to match with the observation. Whereas, as seen in Figure \ref{fig:ws_without_pre_training}, with random initialization it struggles a lot to learn this detail. This means that with pre-training it has learned that cars mostly perform symmetrically on left and right steers, hence it replicates its learning on the other side even if it has seen only left steer on-policy data. It is important to note here that this does not mean it cannot handle steering bias: it can learn to predict that as well from on-policy data, as we will see later in experiments. Also, our results closely compare to APACRace. As APACRace is specifically designed for these scenarios to only adapt friction and aerodynamic parameters for the dynamic bicycle model, it performs slightly better for these scenarios. It is important to note that APACRace assumes availability of geometric and intrinsic vehicle parameters $l_f, l_r, m, I_z, C_{m1}, C_{m2}$.

The numerical results are presented in Table \ref{tab:c5_num}, where we compare our mean lateral error, average speed and number of laps covered. As can be seen, in all metrics, our proposed approach excels when compared to a-e and is close to g.

\begin{table}[]
\vspace{3mm}
\begin{tabular}{|l|l|l|l|}
\hline
                                                                                                               & \begin{tabular}[c]{@{}l@{}}Avg. lateral \\ error (in m)\end{tabular} & \begin{tabular}[c]{@{}l@{}}Avg. speed\\ (in m/s)\end{tabular} & \begin{tabular}[c]{@{}l@{}}Mean No of \\ laps covered\end{tabular} \\ \hline
a) Without adaptation                                                                                             & 1.12                                                                 & 2.12                                                          & 12.3                                                               \\ \hline
\begin{tabular}[c]{@{}l@{}}b) Without adaptation\\ correct params\end{tabular}                                    & 0.94                                                                 & \textbf{2.24}                                                 & 14.2                                                               \\ \hline
\begin{tabular}[c]{@{}l@{}}c) With adaptation\\ random initialization\end{tabular}                                & 1.03                                                                 & 2.04                                                          & 11.6                                                               \\ \hline
\begin{tabular}[c]{@{}l@{}}d) With adaptation\\ + meta-learning\\ + without uncertainty\\ estimation\end{tabular} & 0.33                                                                 & 2.18                                                          & 21.1                                                               \\ \hline
\begin{tabular}[c]{@{}l@{}}e) With adaptation\\ + avg initialization\\ + with uncertainty\\ estimation\end{tabular}    & 0.25                                                        & 2.11                                                          & 22.6                                                      \\ \hline
\begin{tabular}[c]{@{}l@{}}f) With adaptation\\ + meta-learning\\ + with uncertainty\\ estimation\end{tabular}    & 0.22                                                        & 2.18                                                          & \textbf{23.2 }                                                     \\ \hline
\begin{tabular}[c]{@{}l@{}}g) APACRace\end{tabular}    & \textbf{0.14}                                                        & 2.16                                                          & \textbf{23.5}                                                      \\ \hline
\end{tabular}
\caption{Numerical results for numeric sim}
\label{tab:c5_num}
\vspace{-8mm}
\end{table}

\begin{figure}
    \centering
    \includegraphics[width=.45\textwidth]{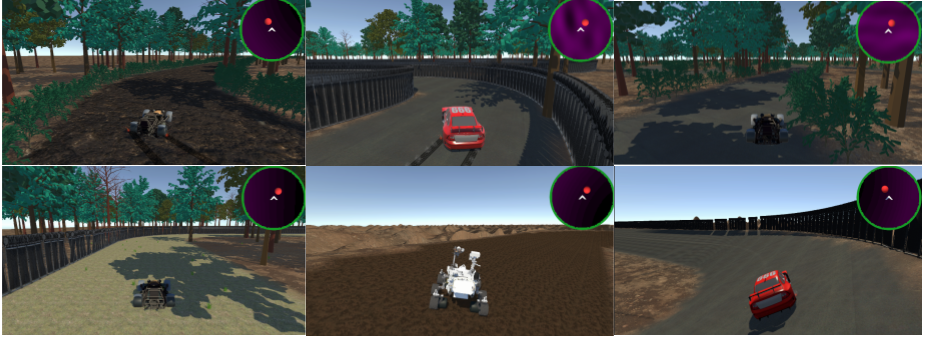}
    \caption{Unity sim environment snapshots}
    \label{fig:unity-envs}
    % \vspace{-mm}
\end{figure}

\subsection{Unity sim} \label{sec:unity_results}

Next, to answer \textbf{Q2}, we test our approach on a custom-built simulator in the Unity game engine on a wide range of scenarios. The comparison in results between our approach and other baselines is given in Table \ref{tab:unity_results}.

The following sim environments are used for testing:

\begin{enumerate}
    \item Circuit track: Tracking a circuit track raceline at control limits involves coordination between longitudinal and lateral control to successfully execute sharp turns.
    \item Steer bias: We add a left steer bias of $0.1$ to the earlier scenario. As APACRace does not assume steer bias, it settles down with a constant lateral error on straights due to steer bias as shown in Figure \ref{fig:unity-steer-bias}, which our proposed approach can handle.
    % FIG HERE
    \begin{figure}[H]
    \vspace{-3mm}
    \centering
        \includegraphics[width=.45\textwidth]{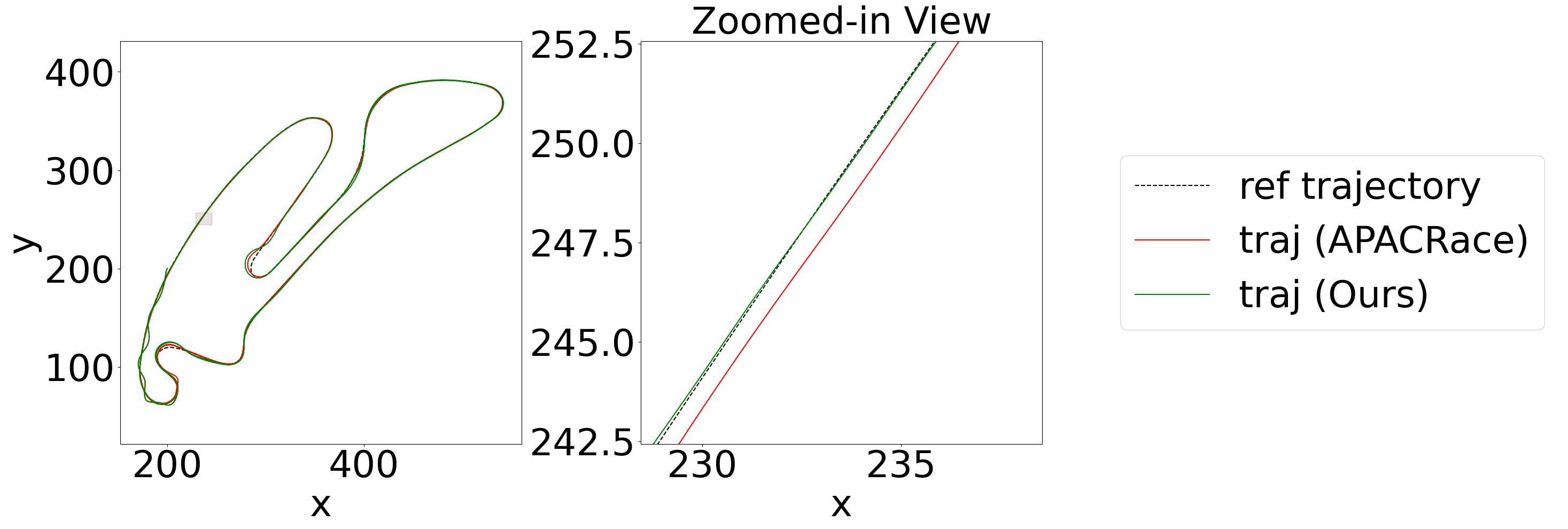}
        \caption{Trajectories followed for Unity steer bias environment}
        \label{fig:unity-steer-bias}
        \vspace{-2.5mm}
    \end{figure}
    \item Decaying friction: We decay the coefficient of friction from $1$ to $0.7$ to test if the model can adapt.
    \item Go-kart: We then change the car to a go-kart, where the rear wheels are coupled by an axle constraining them to move at the same wheel speeds. This results in an opposing torque, which ultimately leads to a different form of dynamics compared to Section \ref{sec:dynamic_model}. Hence, APACRace is not able to handle this and ends up following an unstable trajectory as shown below. %in Figure \ref{fig:unity-kart}.
    % FIG HERE
    \begin{figure}[H]
    \centering
        \includegraphics[width=.45\textwidth]{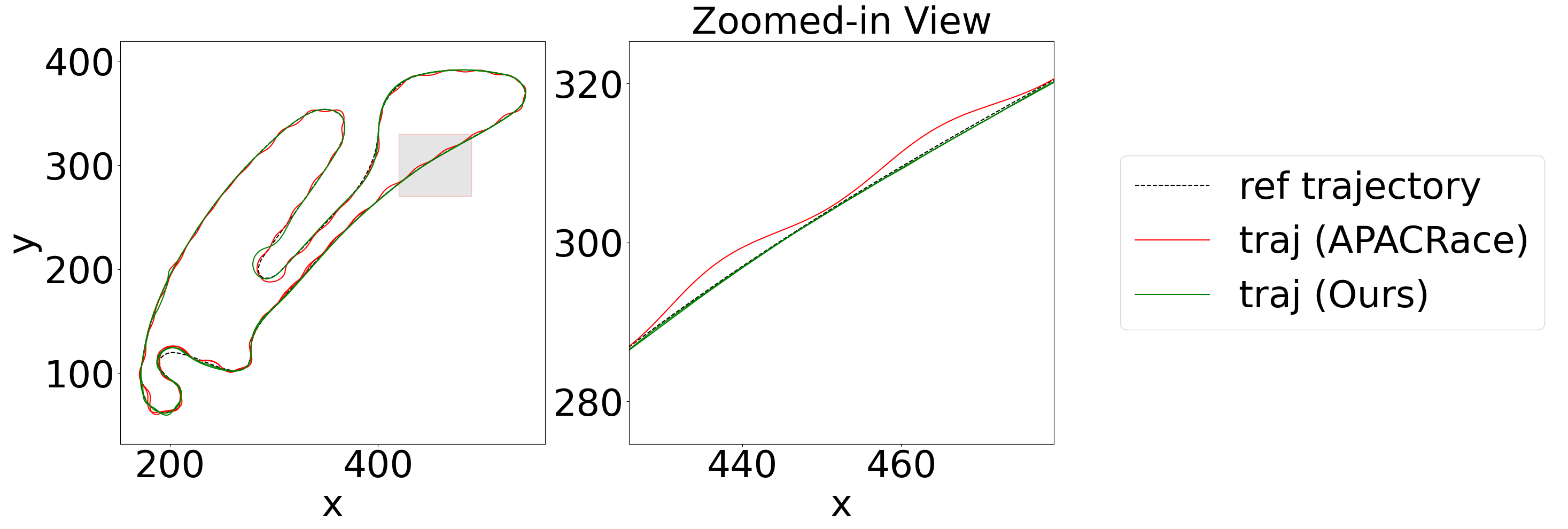}
        \caption{Trajectories followed for Go-Kart environment}
        \label{fig:unity-kart}
        \vspace{-2.5mm}
    \end{figure}
    \item Mud: We then test the approach for a car moving on mud. Mud has very low grip, due to which there is a considerable amount of longitudinal slip. This is ignored in Section \ref{sec:dynamic_model}. It is very challenging to perform in this environment, as it will need to learn to skid in control. APACRace assumes no longitudinal slip, hence performs terribly in this scenario which our approach is able to handle well. We augment longitudinal slip as auxiliary info in the model
    \item Mars: We then test in a different gravity field, which is quite challenging, as it reduces traction. But this can be still learned with model used in APACRace, hence it performs slightly better than our approach, which is completely model-agnostic.
    \item Grass: We then test on a bit of uneven terrain to give it a grassy essence. The uneven terrain can be thought of as exerting forces parallel to the ground plane. 
    This test ensures that the model training is robust to such model noise, which cannot be modeled by our 2D planar model states. 
    % However, this can be enabled to be learned by adding 3D pose information and local heightmap information to our input. But with our 2D planar model state information, the uneven terrain will act as a source of model noise. 
    The results prove that our design is robust to such mild model noise, which does not affect its convergence.
    \item Banked-oval track: We finally test on a track with a bank angle. Moving against an inclined track is particularly challenging to learn, as the vehicle needs to move at an oblique angle towards the slope in order to be able to move straight. Our approach is able to learn to move at an oblique angle in order to be able to stay on the reference, hence performs better than APACRace, which assumes $0$ bank angle.
\end{enumerate}

\begin{table}[]
\vspace{2mm}
\centering
\begin{tabular}{|l|l|l|l|}
\hline
                                                                         & \begin{tabular}[c]{@{}l@{}}Without\\ adaptation\end{tabular} & \begin{tabular}[c]{@{}l@{}}With full model\\ adaptation (ours)\end{tabular} & APACRace            \\ \hline
Circuit                                                                  & 1.31/-                                                       & 0.14/140.1                                                                  & \textbf{0.12/137.2} \\ \hline
\begin{tabular}[c]{@{}l@{}}Circuit with \\ go-kart\end{tabular}          & 0.94/-                                                       & \textbf{0.16/143.6}                                                         & 0.23/154.7          \\ \hline
\begin{tabular}[c]{@{}l@{}}Circuit with \\ steer bias\end{tabular}       & 1.62/-                                                       & \textbf{0.16/142.9}                                                         & 0.38/152.3          \\ \hline
\begin{tabular}[c]{@{}l@{}}Circuit with \\ tire degradation\end{tabular} & 1.99/-                                                       & 0.15/150.2                                                                  & \textbf{0.13/148.9} \\ \hline
Mud                                                                      & 1.43/183.8                                                   & \textbf{0.42/44.7}                                                          & 0.89/75.4           \\ \hline
Mars                                                                     & 1.36/-                                                       & 0.38/70.1                                                                   & \textbf{0.32/66.5}  \\ \hline
Grass                                                                    & 1.29/166.5                                                   & 0.36/45.3                                                                   & \textbf{0.29/42.3}  \\ \hline
Banked-oval                                                              & 1.22/186.8                                                   & \textbf{0.35/43.1}                                                          & 0.53/52.9           \\ \hline
Front-driven                                                             & 1.15/-                                                       & 0.16/139.8                                                                  & \textbf{0.14/137.1} \\ \hline
\end{tabular}
\caption{[Lateral error (in m) / Lap time (in s)] Results table for Unity sim experiments}
\label{tab:unity_results}
\vspace{-2mm}
\end{table}

\subsection{Hardware platform}

We finally also test our approach on a hardware RC car platform to answer \textbf{Q3} and to show that our approach is real-time deployable. We test on an oval track similar in size to what we tested for numeric sim in Section \ref{sec:numeric_sim}. We test on different configurations including towing an object, attaching tape to the wheels, changing surface to alter friction, adding steer bias, and changing the tires to plastic tires. We are able to see consistent results where our approach \textbf{f} is able to beat \textbf{a, d} and \textbf{e}. Also, \textbf{g} performs better for basic scenarios like only changing surface or adding tape to tires. However, with plastic tires, there is a huge tire longitudinal slip where APACRace crashes while our approach is able to perform better. Also, with towed weights and steer bias, our approach works better, as they are not handled by the dynamic bicycle model used by APACRace.

\setlength{\tabcolsep}{4.5pt}
\begin{table}[]
\centering
\begin{tabular}{|l|l|l|l|l|l|}
\hline
                                                                                                               & a & d & e & f & g\\ \hline
1  & 0.81 / 17.2     &   0.21 / 22.5   & 0.18 / 22.7 & 0.13 / 23.2 &  \textbf{0.11 / 23.8}  \\ \hline
  2 & 0.69 / 16.6   &   0.24 / 21.1   & 0.18 / 21.6 & 0.14 / 22.0 &  \textbf{0.13 / 21.9} \\ \hline
  3 & 0.62 / 18.8     &   0.15 / 23.6   & 0.16 / 23.4  & 0.12 / 23.8 &  \textbf{0.09 / 23.8}  \\ \hline
4 & 0.57 / 14.6     &   \textbf{0.13 / 22.5}   & 0.14 / 22.4  & 0.13 / 22.4 &  0.14 / 22.3  \\ \hline
5   & 0.63 / 15.2   &   0.18 / 21.4   & 0.22 / 21.1 & \textbf{0.12 / 22.0} &  0.73 / $11.2^*$   \\ \hline
6   & 0.75 / 16.9   &   0.19 / 22.8   & 0.23 / 21.7 & \textbf{0.13 / 23.4} &  0.17 / 23.4  \\ \hline
\end{tabular}
\caption{Hardware results [lateral error (in m) / avg. lap time (in s)]: * The car crashed into a wall and got stuck, 1: On normal surface, 2: Attached tape to tires, 3: On mat surface, 4: With towed weights, 5: With plastic tires, 6: With steer bias}
\label{tab:c5_hardware}
\vspace{-8mm}
\end{table}

\section{Conclusion} \label{sec:c5_conclusion}

In this work, we saw how meta-learning can be used for an end-to-end trainable LSTM-based vehicle model to allow fast adaptation to any new scenario and changes within the scenario online. We saw how using epistemic uncertainty estimation and an uncertainty-aware cost function for MPPI further improves the tracking performance. The same structure can be extended to other robots like quadrupeds, drones, manipulators, etc. which we would like to explore in future.

\bibliographystyle{IEEEtran}
\bibliography{./IEEEfull,refs}

\begin{thebibliography}{10}
\providecommand{\url}[1]{#1}
\csname url@rmstyle\endcsname
\providecommand{\newblock}{\relax}
\providecommand{\bibinfo}[2]{#2}
\providecommand\BIBentrySTDinterwordspacing{\spaceskip=0pt\relax}
\providecommand\BIBentryALTinterwordstretchfactor{4}
\providecommand\BIBentryALTinterwordspacing{\spaceskip=\fontdimen2\font plus
\BIBentryALTinterwordstretchfactor\fontdimen3\font minus \fontdimen4\font\relax}
\providecommand\BIBforeignlanguage[2]{{%
\expandafter\ifx\csname l@#1\endcsname\relax
\typeout{** WARNING: IEEEtran.bst: No hyphenation pattern has been}%
\typeout{** loaded for the language `#1'. Using the pattern for}%
\typeout{** the default language instead.}%
\else
\language=\csname l@#1\endcsname
\fi
#2}}

\bibitem{Kaufmann2023ChampionlevelDR}
\BIBentryALTinterwordspacing
E.~Kaufmann, L.~Bauersfeld, A.~Loquercio, M.~M{\"u}ller, V.~Koltun, and D.~Scaramuzza, ``Champion-level drone racing using deep reinforcement learning,'' \emph{Nature}, vol. 620, pp. 982 -- 987, 2023. [Online]. Available: \url{https://api.semanticscholar.org/CorpusID:261357832}
\BIBentrySTDinterwordspacing

\bibitem{Wischnewski2022IndyAC}
\BIBentryALTinterwordspacing
A.~Wischnewski, M.~Geisslinger, J.~Betz, T.~Betz, F.~Fent, A.~Heilmeier, L.~Hermansdorfer, T.~Herrmann, S.~Huch, P.~Karle, F.~Nobis, L.~Ogretmen, M.~Rowold, F.~Sauerbeck, T.~Stahl, R.~Trauth, M.~Lienkamp, and B.~Lohmann, ``Indy autonomous challenge - autonomous race cars at the handling limits,'' \emph{ArXiv}, vol. abs/2202.03807, 2022. [Online]. Available: \url{https://api.semanticscholar.org/CorpusID:246652188}
\BIBentrySTDinterwordspacing

\bibitem{Senanayake2024TheRO}
\BIBentryALTinterwordspacing
R.~Senanayake, ``The role of predictive uncertainty and diversity in embodied ai and robot learning,'' \emph{ArXiv}, vol. abs/2405.03164, 2024. [Online]. Available: \url{https://api.semanticscholar.org/CorpusID:269605587}
\BIBentrySTDinterwordspacing

\bibitem{Chen2021UnderstandingDR}
\BIBentryALTinterwordspacing
X.~Chen, J.~Hu, C.~Jin, L.~Li, and L.~Wang, ``Understanding domain randomization for sim-to-real transfer,'' \emph{ArXiv}, vol. abs/2110.03239, 2021. [Online]. Available: \url{https://api.semanticscholar.org/CorpusID:242784472}
\BIBentrySTDinterwordspacing

\bibitem{Kumar2021RMARM}
A.~Kumar, Z.~Fu, D.~Pathak, and J.~Malik, ``Rma: Rapid motor adaptation for legged robots,'' \emph{ArXiv}, vol. abs/2107.04034, 2021.

\bibitem{Xiao2023SafeDP}
\BIBentryALTinterwordspacing
W.~Xiao, T.~He, J.~Dolan, and G.~Shi, ``Safe deep policy adaptation,'' \emph{ArXiv}, vol. abs/2310.08602, 2023. [Online]. Available: \url{https://api.semanticscholar.org/CorpusID:264128072}
\BIBentrySTDinterwordspacing

\bibitem{Liniger2015OptimizationbasedAR}
\BIBentryALTinterwordspacing
A.~Liniger, A.~Domahidi, and M.~Morari, ``Optimization‐based autonomous racing of 1:43 scale rc cars,'' \emph{Optimal Control Applications and Methods}, vol.~36, pp. 628 -- 647, 2015. [Online]. Available: \url{https://api.semanticscholar.org/CorpusID:11242645}
\BIBentrySTDinterwordspacing

\bibitem{Williams2016AggressiveDW}
\BIBentryALTinterwordspacing
G.~Williams, P.~Drews, B.~Goldfain, J.~M. Rehg, and E.~A. Theodorou, ``Aggressive driving with model predictive path integral control,'' \emph{2016 IEEE International Conference on Robotics and Automation (ICRA)}, pp. 1433--1440, 2016. [Online]. Available: \url{https://api.semanticscholar.org/CorpusID:18322548}
\BIBentrySTDinterwordspacing

\bibitem{Chrosniak2023DeepDV}
\BIBentryALTinterwordspacing
J.~Chrosniak, J.~Ning, and M.~Behl, ``Deep dynamics: Vehicle dynamics modeling with a physics-informed neural network for autonomous racing,'' \emph{ArXiv}, vol. abs/2312.04374, 2023. [Online]. Available: \url{https://api.semanticscholar.org/CorpusID:266044227}
\BIBentrySTDinterwordspacing

\bibitem{Kalaria2023AdaptivePA}
\BIBentryALTinterwordspacing
D.~Kalaria, Q.~Lin, and J.~M. Dolan, ``Adaptive planning and control with time-varying tire models for autonomous racing using extreme learning machine,'' \emph{ArXiv}, vol. abs/2303.08235, 2023. [Online]. Available: \url{https://api.semanticscholar.org/CorpusID:257532643}
\BIBentrySTDinterwordspacing

\bibitem{Nagabandi2018LearningTA}
\BIBentryALTinterwordspacing
A.~Nagabandi, I.~Clavera, S.~Liu, R.~S. Fearing, P.~Abbeel, S.~Levine, and C.~Finn, ``Learning to adapt in dynamic, real-world environments through meta-reinforcement learning,'' \emph{arXiv: Learning}, 2018. [Online]. Available: \url{https://api.semanticscholar.org/CorpusID:56475856}
\BIBentrySTDinterwordspacing

\bibitem{Kabzan2019AMZDT}
J.~Kabzan, M.~de~la Iglesia~Valls, V.~Reijgwart, \emph{et~al.}, ``{AMZ} driverless: The full autonomous racing system,'' \emph{Journal of Field Robotics}, vol.~37, pp. 1267 -- 1294, 2019.

\bibitem{Kalaria2022DelayAwareRC}
\BIBentryALTinterwordspacing
D.~Kalaria, Q.~Lin, and J.~M. Dolan, ``Delay-aware robust control for safe autonomous driving and racing,'' \emph{IEEE Transactions on Intelligent Transportation Systems}, vol.~25, pp. 7140--7150, 2022. [Online]. Available: \url{https://api.semanticscholar.org/CorpusID:260499046}
\BIBentrySTDinterwordspacing

\bibitem{Kalaria2021LocalNO}
D.~Kalaria, P.~Maheshwari, A.~Jha, A.~K. Issar, D.~Chakravarty, S.~Anwar, and A.~Towar, ``Local {NMPC} on global optimised path for autonomous racing,'' \emph{ArXiv}, vol. abs/2109.07105, 2021.

\bibitem{Pacejka1982TireAV}
H.~B. Pacejka, ``Tire and vehicle dynamics.''\hskip 1em plus 0.5em minus 0.4em\relax Elsevier, 2005.

\bibitem{Finn2017ModelAgnosticMF}
\BIBentryALTinterwordspacing
C.~Finn, P.~Abbeel, and S.~Levine, ``Model-agnostic meta-learning for fast adaptation of deep networks,'' in \emph{International Conference on Machine Learning}, 2017. [Online]. Available: \url{https://api.semanticscholar.org/CorpusID:6719686}
\BIBentrySTDinterwordspacing

\bibitem{Williams2018RobustSB}
\BIBentryALTinterwordspacing
G.~Williams, B.~Goldfain, P.~Drews, K.~Saigol, J.~M. Rehg, and E.~A. Theodorou, ``Robust sampling based model predictive control with sparse objective information,'' \emph{Robotics: Science and Systems XIV}, 2018. [Online]. Available: \url{https://api.semanticscholar.org/CorpusID:46964005}
\BIBentrySTDinterwordspacing

\bibitem{Yi2024CoVOMPCTA}
\BIBentryALTinterwordspacing
Z.~Yi, C.~Pan, G.~He, G.~Qu, and G.~Shi, ``Covo-mpc: Theoretical analysis of sampling-based mpc and optimal covariance design,'' \emph{ArXiv}, vol. abs/2401.07369, 2024. [Online]. Available: \url{https://api.semanticscholar.org/CorpusID:266999348}
\BIBentrySTDinterwordspacing

\bibitem{Yin2022RiskAwareMP}
\BIBentryALTinterwordspacing
J.~Yin, Z.~Zhang, and P.~Tsiotras, ``Risk-aware model predictive path integral control using conditional value-at-risk,'' \emph{2023 IEEE International Conference on Robotics and Automation (ICRA)}, pp. 7937--7943, 2022. [Online]. Available: \url{https://api.semanticscholar.org/CorpusID:252532183}
\BIBentrySTDinterwordspacing

\end{thebibliography}

% \appendix

% \begin{figure}
% \centering
% \begin{subfigure}{.23\textwidth}
%     \centering
%     \includegraphics[width=\textwidth]{c5/rc/vicon.png}
%     \caption{On normal surface}
%     \label{fig:c5_rc_norm}
% \end{subfigure}
% \begin{subfigure}{.23\textwidth}
%     \centering
%     \includegraphics[width=\textwidth]{c5/rc/vicon2.png}
%     \caption{Attaching tape to tires}
%     \label{fig:c5_rc_tape}
% \end{subfigure}
% \begin{subfigure}{.23\textwidth}
%     \centering
%     \includegraphics[width=\textwidth]{c5/rc/vicon3.png}
%     \caption{On mat surface}
%     \label{fig:c5_rc_mat}
% \end{subfigure}
% \begin{subfigure}{.23\textwidth}
%     \centering
%     \includegraphics[width=\textwidth]{c5/rc/vicon1.png}
%     \caption{With towed weights}
%     \label{fig:c5_rc_wt}
% \end{subfigure}
% \caption{Trajectory followed by RC car with our approach}
% \label{fig:rc}
% \end{figure}
\end{document}